\documentclass[10pt,twocolumn,letterpaper]{article}

\usepackage[pagenumbers]{cvpr} 

\usepackage{algorithm}
\usepackage{algorithmic}
\usepackage{etoc}
\usepackage{times}
\usepackage{epsfig}
\usepackage{graphicx}
\usepackage{amsmath}
\usepackage{amssymb}
\usepackage{amsfonts}
\usepackage{mathtools}
\usepackage{bm}           
\usepackage{microtype}

\usepackage{booktabs}     
\usepackage{multirow}     
\usepackage{multicol}     
\usepackage{array}        
\usepackage{tabularx}     

\usepackage{xcolor}

\newcommand{\nerfify}{\textsc{Nerfify}}
\newcommand{\nerfifybench}{\textsc{Nerfify-Bench}}

\definecolor{codegreen}{rgb}{0,0.6,0}
\definecolor{codegray}{rgb}{0.5,0.5,0.5}
\definecolor{codepurple}{rgb}{0.58,0,0.82}
\definecolor{backcolour}{rgb}{0.95,0.95,0.92}
\definecolor{deepblue}{rgb}{0,0,0.5}
\definecolor{deepred}{rgb}{0.6,0,0}
\definecolor{deepgreen}{rgb}{0,0.5,0}

\definecolor{excellentgreen}{RGB}{144,238,144}
\definecolor{goodgreen}{RGB}{200,255,200}
\definecolor{warningyellow}{RGB}{255,255,153}
\definecolor{badred}{RGB}{255,200,200}

\usepackage{listings}

\lstdefinestyle{pythonstyle}{
    language=Python,
    backgroundcolor=\color{backcolour},   
    commentstyle=\color{codegreen},
    keywordstyle=\color{deepblue}\bfseries,
    numberstyle=\tiny\color{codegray},
    stringstyle=\color{deepred},
    basicstyle=\ttfamily\scriptsize,
    breakatwhitespace=false,         
    breaklines=true,                 
    captionpos=b,                    
    keepspaces=true,                 
    numbers=left,                    
    numbersep=5pt,                  
    showspaces=false,                
    showstringspaces=false,
    showtabs=false,                  
    tabsize=2,
    frame=single,
    rulecolor=\color{black},
    xleftmargin=2em,
    framexleftmargin=1.5em
}

\lstdefinestyle{compactstyle}{
    language=Python,
    backgroundcolor=\color{backcolour},
    commentstyle=\color{codegreen},
    keywordstyle=\color{deepblue},
    stringstyle=\color{deepred},
    basicstyle=\ttfamily\tiny,
    breaklines=true,
    captionpos=b,
    showstringspaces=false,
    frame=single,
    numbers=none
}
\usepackage[table]{xcolor}
\usepackage{colortbl}
\usepackage{colortbl}     

\usepackage{microtype}    
\usepackage{enumitem}     
\usepackage{caption}      
\usepackage{subcaption}   

\usepackage{pifont}       

\usepackage{url}          
\usepackage{comment}      

\definecolor{cvprblue}{rgb}{0.21,0.49,0.74}
\usepackage[pagebackref,breaklinks,colorlinks,allcolors=cvprblue]{hyperref}

\def\paperID{} 
\def\confName{CVPR}
\def\confYear{2026}

\title{\nerfify: A Multi-Agent Framework for Turning NeRF Papers into Code}

\author{Seemandhar Jain \quad Keshav Gupta \quad Kunal Gupta \quad Manmohan Chandraker\\
University of California, San Diego\\
{\tt\small \{sejain, keg019, k5gupta, mkchandraker\}@ucsd.edu}
}

\begin{document}
\maketitle
\begin{abstract}

The proliferation of neural radiance field (NeRF) research requires significant efforts to reimplement papers before building upon them. We introduce \nerfify, a multi-agent framework that reliably converts NeRF research papers into trainable Nerfstudio plugins, in contrast to generic paper-to-code methods and frontier models like GPT-5 that usually fail to produce runnable code. \nerfify\ achieves domain-specific executability through six key innovations: (1) Context-free grammar (CFG): LLM synthesis is constrained by Nerfstudio formalized as a CFG, ensuring generated code satisfies architectural invariants. (2) Graph-of-Thought code synthesis: Specialized multi-file-agents generate repositories in topological dependency order, validating contracts and errors at each node. (3) Compositional citation recovery: Agents automatically retrieve and integrate components (samplers, encoders, proposal networks) from citation graphs of references. (4) Visual feedback: Artifacts are diagnosed through PSNR-minima ROI analysis, cross-view geometric validation, and VLM-guided patching to iteratively improve quality. (5) Knowledge enhancement: Beyond reproduction, methods can be improved with novel optimizations. (6) Benchmarking: An evaluation framework is designed for NeRF paper-to-code synthesis across 30 diverse papers. On papers without public implementations, \nerfify\ achieves visual quality matching expert human code (±0.5 dB PSNR, ±0.2 SSIM) while reducing implementation time from weeks to minutes. \nerfify\ demonstrates that a domain-aware design enables code translation for complex vision papers, potentiating accelerated and democratized reproducible research. Code, data and implementations will be publicly released.

\end{abstract}    
\section{Introduction}
\label{sec:intro}

\begin{figure}[t]
\centering
\includegraphics[width=\linewidth]{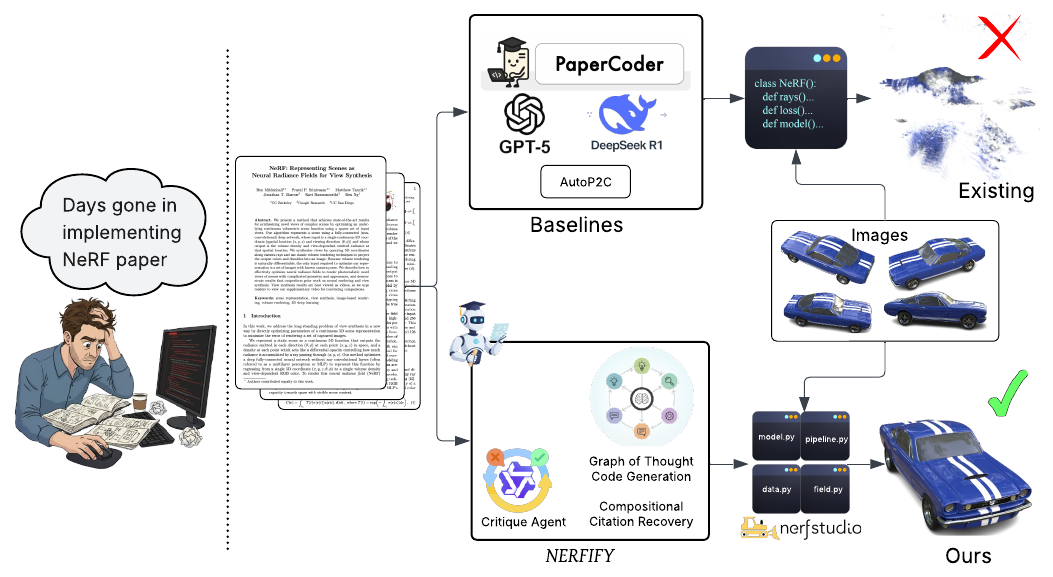}
\vspace{-0.7cm}
\caption{\textbf{Overview.} Manual NeRF implementation requires weeks of specialized effort (left). Existing paper-to-code systems fail to produce trainable code. \nerfify\ automates this process through grammar-constrained synthesis and compositional citation recovery, generating fully trainable Nerfstudio plugins in minutes (right).}
\vspace{-0.35cm}
\label{fig:teaser}
\end{figure}

Since its publication in 2020, the Neural Radiance Fields (NeRF) paper \cite{mildenhall2020nerf} has spawned over 1,000 follow-ups. Yet, unavailability of code or standardized implementations for most of them \cite{wang2025nerfArxivDaily} means each subsequent work requires a significant effort to reimplement existing NeRF papers. While the issue exists in broader machine learning research \cite{raff2019step}, effective NeRF implementations present unique challenges. We consider the question of devising a large language model (LLM) agent to automatically produce trainable, performant and standardized NeRFs from their papers.

\vspace{0.07cm}
\noindent \textbf{NeRF implementations are uniquely challenging,} with expertise required across volumetric rendering, computer vision and neural optimization. A single misplaced activation or incorrect ray-sphere intersection produces failures ranging from the catastrophic (NaN gradients) to the subtle (degenerate solutions). Debugging requires understanding whether failures stem from code bugs, scene geometry, hyperparameters, or the paper's own ambiguities, with computationally demanding cycles (the original NeRF requires 100k-300k iterations taking 24-48 hours on high-end GPUs \cite{mildenhall2020nerf}).

\vspace{0.07cm}
\noindent \textbf{Generic paper-to-code approaches do not suffice,} given that the current best-performing system, O1, achieves only 26.6\% accuracy on complex papers compared to 41.4\% for human experts \cite{starace2025paperbench}. The challenge is exacerbated for NeRFs, where the strong domain coupling needed for its unique combination of rendering mathematics and neural architectures leads to execution failures and large performance gaps. As discussed in Section~\ref{proposed} and exhaustively shown in the Supplementary, recent advances like Paper2Code \cite{seo2025paper2code} and AutoP2C \cite{lin2025autop2c} -- despite progress on general machine learning tasks -- usually do not produce executable code or trainable NeRFs, with issues ranging from incorrect implementations (such as K-Planes with regular MLPs instead of planar factorization), or catastrophic failures (producing only dataset loading code). Another key challenge is that modern NeRF papers build upon chains of dependencies that generic systems cannot resolve, for example, see Figure \ref{fig:nerf_dependency_graph} for K-Planes~\cite{fridovich2023k}. A single phrase such as ``we adopt the distortion loss from \cite{barron2022mipnerf360}'’ requires navigating to that paper, locating the correct equation, translating to executable code and implementing stop-gradient operations critical for stable training. 

\vspace{0.08cm}
\noindent \textbf{We introduce \nerfify, a multi-agent framework} that reliably converts NeRF research papers into code that trains, converges and matches the visual quality of expert implementations (Figure \ref{fig:teaser}). Its success is based on five key technical innovations. 1) Use of Nerfstudio \cite{tancik2023nerfstudio} architecture as a context-free grammar (CFG) that encodes domain-aware module compositions and interface contracts, leading to  LLM synthesis of architecturally correct code by construction. 2) Graph-of-thought code generation to coordinate specialized agents that generate multi-file repositories in topological order, validating type signatures, tensor shapes and circular dependencies at each dependency before proceeding. 3) A compositional citation recovery that traverses reference graphs to retrieve implicit dependencies (such as proposal networks from Mip-NeRF 360 \cite{barron2022mipnerf360}, hash encoders from Instant-NGP \cite{muller2022instant}). 4) A critique agent that provides feedback by analyzing training runs to diagnose visual artifacts through PSNR-minima analysis and cross-view geometric validation, using vision-language models (VLMs) to guide targeted fixes. 5) Besides reproduction, an option to improve with optimizations where applicable.

\vspace{0.08cm}
\noindent \textbf{We extensively evaluate with \nerfifybench}, a novel benchmark with 30 diverse papers stratified across code status and implementation complexity. \nerfify\ achieves full executability and rendering quality within 0.5 dB PSNR of expert implementations, while generic baselines fail to produce trainable code in 95\% of cases. 

\vspace{0.08cm}
\noindent \textbf{\nerfify\ will accelerate reproducible research,} allowing researchers to reproduce complex NeRF papers in hours rather than weeks, rapidly prototype techniques and make accessible to the community papers that would otherwise remain purely theoretical. As the NeRF community continues its growth, \nerfify\ ensures that every paper's contributions become available to all researchers, democratizing access to cutting-edge techniques and enabling the compositional research that drives the field forward. With our demonstration that depth and specialization in a multi-agent framework unlocks transformative capabilities for NeRFs, we believe future work will achieve similar transformations for other communities too. Code and data for \nerfify, \nerfifybench\ and generated implementations will be publicly released.

\section{Related Work}
\label{sec:related_work}


\textbf{Generic Paper-to-Code Systems.}
\label{subsec:paper2code}
Paper2Code~\cite{seo2025paper2code} employs a three-stage pipeline (planning, analysis, generation), but its generic design lacks neural field architectures for NeRF implementation. AutoP2C~\cite{lin2025autop2c} advances multimodal understanding with high executability on recent papers, but generates non trainable repositories with just placeholders for NeRFs. AutoReproduce~\cite{zhao2025autoreproduce} introduces paper lineage-extracting domain knowledge from citations.
On PaperBench~\cite{openai2024paperbench}, Claude 3.5 Sonnet achieves only 21\% accuracy versus 41.4\% for human researchers on ICML papers, motivating our domain-specific approach. RPG~\cite{luo2025rpg} employs test-driven development with repository planning graphs but cannot effectively parse papers or extract mathematical formulations. Paper2Agent~\cite{miao2025paper2agent} converts papers into conversational interfaces rather than trainable implementations. Recent works \cite{zhou2025reflective,gandhi2025researchcodeagent} explore fine-grained verification and dynamic planning, but remain domain-agnostic.

\vspace{0.1cm}
\noindent \textbf{Multi-Agent Code Generation}
has emerged as a key field for complex code synthesis \cite{hong2023metagpt, qian2023chatdev, wu2024autogen, huang2024agentcoder},
with recent systems proposing repository-level code generation \cite{wang2024codeact, zhang2024autocoderover, cognition2024devin, wang2024opendevin, yang2024swe, holt2023l2mac}. However, general software systems lack the mathematical understanding and architectural constraints required for successful synthesis of research code.

\vspace{0.1cm}
\noindent \textbf{Domain-Specific Code Synthesis} outperforms generic approaches by encoding specialized knowledge. Scene Language \cite{zhang2025scene} uses CFG to structure visual program synthesis, CODEP~\cite{wang2023codep} employs Pushdown Automaton and TSL+LLM \cite{mavrogiannis2024tsl} combines Temporal Stream Logic to act as CFG.

\vspace{0.1cm}
\noindent \textbf{Feedback and Planning Approaches} like Graph of Thoughts (GoT) \cite{besta2024got} generalize reasoning into directed graphs enabling aggregation, refinement loops, and backtracking, while Tree of Thoughts \cite{yao2023tot} and Tree-of-Code \cite{ni2024tree} explore over reasoning trees. Self-Debugging \cite{chen2023teaching}, Reflexion \cite{shinn2023reflexion}, Self-Refine \cite{madaan2023selfrefine}, Clover \cite{sun2024clover} and xKG \cite{luo2025executable} introduce refinement and consistency checks, but either require reference databases or lack domain-specific constraints.

\section{Method}

\begin{figure*}[t]
  \centering
  \includegraphics[width=\textwidth]{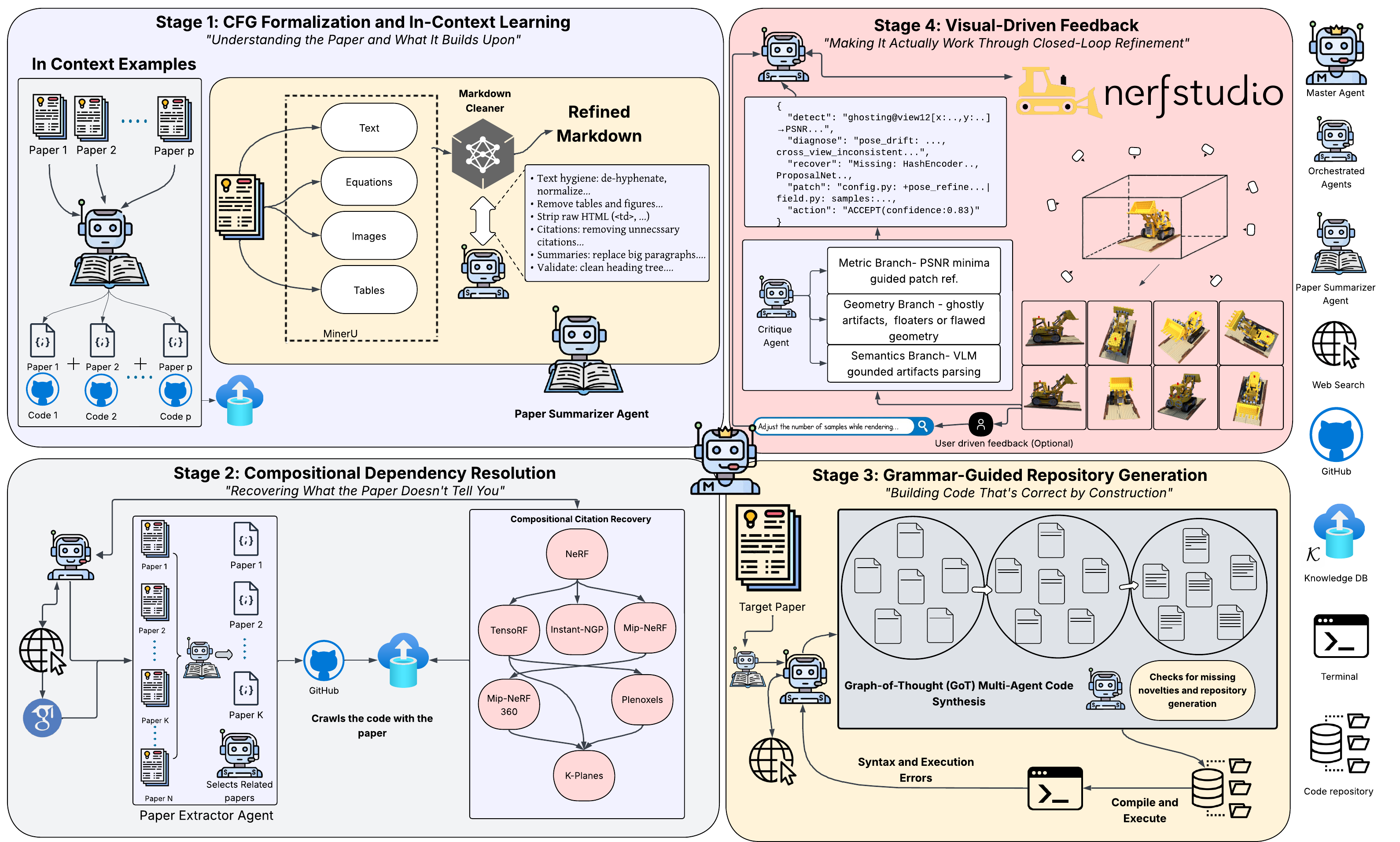}
  \caption{\nerfify\ converts NeRF papers into code through four stages: (1) Agent parses and summarizes PDFs into simple markdown, CFG from Nerfstudio and curated paper-code pairs as In-Context examples are saved in $\mathcal{K}$ (2) Compositional dependency resolution traverses citation graphs to retrieve missing components from referenced papers, (3) GoT code synthesis generates repository files through specialized agents operating in topological order (4) Visual refinement iteratively patches artifacts until achieving expert-level quality.}
  \label{proposed}
\end{figure*}


\subsection{Problem Statement}
Given a NeRF research paper $\mathcal{P}$, our objective is to synthesize an executable repository $\mathcal{C}$ that faithfully implements the described method within the Nerfstudio framework. We formalize this paper-to-code synthesis problem as follows.
\noindent\textbf{Repository Definition.}
A repository $\mathcal{C}$ consists of a set of files and their dependency structure:
\begin{equation}
    \mathcal{C} = (F, G), \quad F = \{f_1, f_2, \ldots, f_n\}
\end{equation}
where $G = \text{BuildRepoDAG}(F)$ is a directed acyclic graph with vertices $V(G) = F$ representing files and edges $E(G) \subseteq F \times F$ encoding import and dataflow dependencies. The acyclic constraint ensures compilability: $(f_i, f_j) \in E(G) \implies$ no path exists from $f_j$ to $f_i$.

\noindent\textbf{Paper Representation}
We extract structured information from paper $\mathcal{P}$ through a comprehensive parsing pipeline:
\begin{align}
    \mathcal{E}(\mathcal{P}) &= \langle T(\mathcal{P}), I(\mathcal{P}), Q(\mathcal{P}), B(\mathcal{P}) \rangle, \\
    T(\mathcal{P}) &= \langle H, \{p_i\}_{i=1}^{n_p}, \{a_\ell\}_{\ell=1}^{n_a}, \{c_k\}_{k=1}^{n_c}, \{r_m\}_{m=1}^{n_r} \rangle.
\end{align}
The textual component $T(\mathcal{P})$ encompasses section headings $H$, paragraphs $\{p_i\}$, algorithm blocks $\{a_\ell\}$, figure captions $\{c_k\}$, and references $\{r_m\}$. Visual content $I(\mathcal{P})$ includes architectural diagrams, figures, and result visualizations that often contain crucial implementation details not present in the text. The mathematical component $Q(\mathcal{P})$ captures equations and formulas that define the core algorithms, while bibliographic metadata $B(\mathcal{P})$ provides citation information essential for compositional recovery.

\noindent\textbf{Agent Formulation}
Let $\mathcal{A}$ be a multi-agent system that maps paper representations to executable code using auxiliary resources:
$\mathcal{A}: (\mathcal{E}(\mathcal{P}); \mathcal{R}) \mapsto \mathcal{C}$.
The resource tuple $\mathcal{R} = (\mathcal{K}, \mathcal{W}, \mathcal{X})$ comprises three essential components. The domain knowledge base $\mathcal{K}$ encodes NeRF-specific architectural patterns, common implementation strategies, and framework conventions (CFG) accumulated from analyzing existing implementations. Web resources $\mathcal{W}$ enable dynamic retrieval of cited papers, missing components, and implementation details referenced but not fully specified in the target paper. Finally, the repository $\mathcal{X}$ provides code templates and reference implementations (in-context examples) that serve as bases for synthesis.


\subsection{Proposed Multi-Stage Framework}
\nerfify\ has four stages that transform research papers into executable code. Each stage addresses specific challenges in automated code synthesis for complex vision systems, from paper understanding through visual quality refinement.


\subsubsection*{Stage 1: CFG Formalization and In-Context Learning}
A foundation of \nerfify\ is constructing the domain knowledge base $\mathcal{K}$ by formalizing Nerfstudio's architectural patterns as a context-free grammar (CFG) that constrains code generation. We curate pairs $\{(\mathcal{P}_i, \mathcal{C}_i)\}_{i=1}^m$ of NeRF papers and their corresponding implementations, which populate $\mathcal{K}$ and serve as in-context examples $\mathcal{X}$ for our synthesis pipeline. The extraction function $\mathcal{E}(\cdot)$ employs MinerU~\cite{niu2025mineru2}, a state-of-the-art PDF-to-markdown conversion tool.

As shown in Figure \ref{proposed}, the paper summarization agent processes each document through multiple refinement stages. First, MinerU converts the PDF into a structured markdown representation preserving equations, tables, figures, references and implementation details. A subsequent cleaning agent then distills this raw conversion by removing irrelevant sections such as extended introductions, related work discussions and redundant references. This agent operates under strict preservation constraints: all equations, implementation pseudocode, architectural diagrams, and citation relationships must remain intact. The agent validates completeness by ensuring that key technical components identified in the abstract appear in the refined document. Each processed paper and its corresponding Nerfstudio implementation are then stored as structured examples in our knowledge base, forming the grammatical foundation that guides subsequent synthesis. Full details of the CFG formalization are provided in the supplementary material.


\subsubsection*{Stage 2: Compositional Dependency Resolution}
NeRF papers are inherently compositional -- 
for instance, ZipNeRF implementation requires the proposal network from Mip-NeRF 360, hash encoding from Instant-NGP, and anti-aliasing techniques that span three papers. Na\"{i}ve approaches that retrieve only the target paper fail because critical implementation details reside in dependencies. Our paper extractor agent constructs a citation dependency graph $G' = (V', E')$ of all referenced papers and their transitive dependencies, 
with nodes $v \in V'$ representing papers and edges $(u, v) \in E'$ indicating that $v$ depends on technical components from $u$. Given a target paper $P_{\text{target}}$, our agent performs iterative multi-hop retrieval through four steps:

\vspace{0.1cm}
\noindent {\em 1. Dependency discovery:} Parse $P_{\text{target}}$ to extract cited papers $\mathcal{C} = \{c_1, \ldots, c_n\}$ and identify which components are borrowed (e.g., ``we use the proposal network from~\cite{barron2022mipnerf360}'').

\vspace{0.1cm}
\noindent {\em 2. Recursive resolution:} For each $c_i \in \mathcal{C}$, retrieve: 
$$\text{Dependencies}(c_i) = \{c_i\} \cup \bigcup_{d \in \text{cited}(c_i)} \text{Dependencies}(d),$$
where the recursion terminates when all components required to implement the target paper have been located.

\vspace{0.1cm}
\noindent {\em 3. Component extraction:} For each paper in the transitive closure, extract specific components mentioned in $P_{\text{target}}$ or required by intermediate dependencies. Specialized LLM agents identify (a) architectural modules: e.g., ``hash encoding'', (b) loss functions: e.g., ``distortion loss'', (c) training protocols: e.g., ``proposal sampling with stop-gradient''.

\vspace{0.1cm}
\noindent {\em 4. Termination criterion:} Stop when the agent has retrieved all components needed to satisfy the interface contracts of $P_{\text{target}}$ and no unresolved dependencies remain in the graph.

\begin{figure}[t]
\centering
\includegraphics[width=\columnwidth]{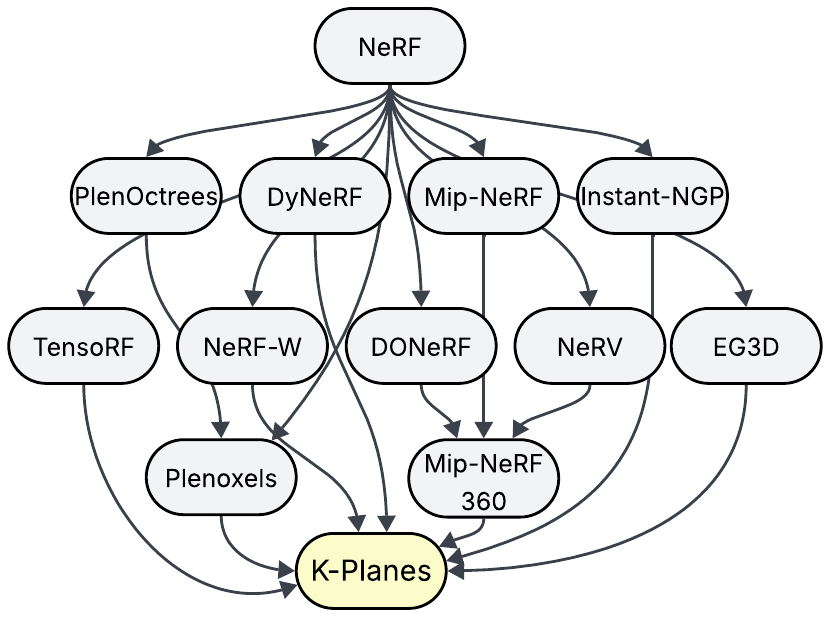}
\caption{\textbf{NeRF citation dependency graphs.} Implementing K-Planes requires retrieving components from 7 direct dependencies (Plenoxels, TensoRF, Instant-NGP, Mip-NeRF 360, DyNeRF, EG3D, NeRF-W) and 12 total papers with transitive dependencies. Our compositional citation recovery automatically traverses such graphs to identify and retrieve all necessary components.}
\label{fig:nerf_dependency_graph}
\end{figure}

\vspace{0.1cm}
Figure~\ref{fig:nerf_dependency_graph} illustrates this process through K-Planes, which depends on 7 different papers. Our agent: \\
{\em 1. Identifies direct citations:} Plenoxels (optimization), TensoRF (factorization), Instant-NGP (hash grids), Mip-NeRF 360 (proposal networks), DyNeRF (temporal sampling), EG3D (triplanes), NeRF-W (appearance codes).\\
{\em 2. Recursively retrieves dependencies:} Mip-NeRF 360 requires Mip-NeRF, DONeRF, and NeRV; Plenoxels requires PlenOctrees; each ultimately traces back to NeRF. \\
{\em 3. Extracts components:} proposal network and distortion loss from Mip-NeRF 360, hash encoder from Instant-NGP and VM decomposition from TensoRF. \\
{\em 4. Terminates:} when all papers are processed and K-Planes' planar factorization $f(\mathbf{q}) = \prod_{c \in \mathcal{C}} f(\mathbf{q})_c$ can be implemented using the retrieved components.

\begin{figure*}[t]
  \centering
  \includegraphics[width=\linewidth]{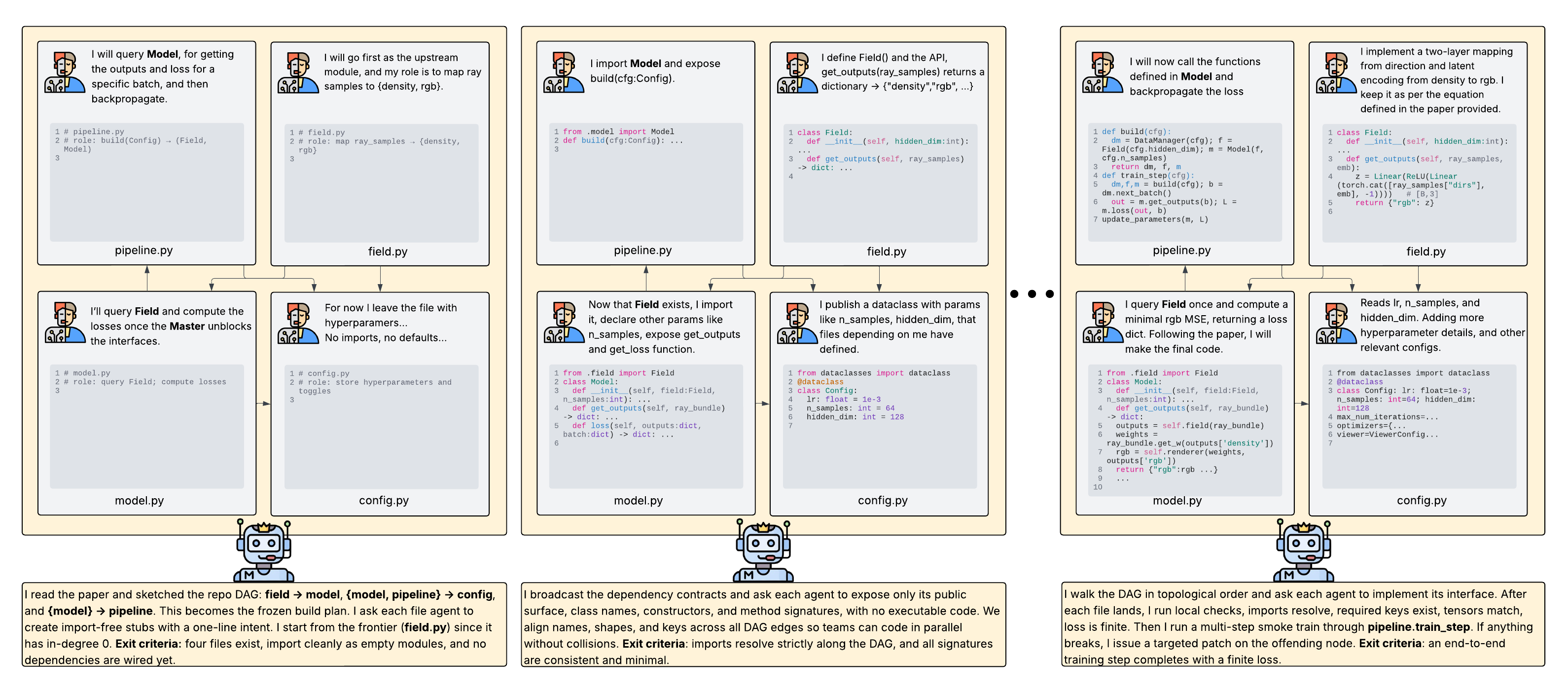}
  \caption{\textbf{Graph-of-Thought (GoT) Multi-Agent Code Synthesis}. The master agent orchestrates specialized file-agents that progressively build a NeRF repository over k steps. Each step shows files being created or modified through four stages: (1) DAG Construction maps papers to Nerfstudio component dependencies, (2) Interface Freeze establishes API contracts in topological order, (3) Implementation generates validated code with shape/gradient checks, (4) Integration Testing runs smoke tests with automated repair. Files evolve from minimal interfaces to complete implementations as agents coordinate through the dependency graph, producing runnable NeRF plugins.}
  \label{fig:got-kstep}
\end{figure*}

\subsubsection*{Stage 3: Grammar-Guided Repository Generation}

The master synthesis agent
orchestrates code generation through a Graph-of-Thought (GoT) approach that ensures trainable code through syntax and execution feedback loops. It captures the tight inter-file couplings inherent in NeRF pipelines, where configurations flow to data managers, which feed fields, models, and ultimately the training pipeline.

As illustrated in Figure~\ref{fig:got-kstep}, the GoT code synthesis proceeds in four phases. During {\em DAG construction}, the parsed paper is mapped to a dependency graph over core Nerfstudio components. {\em Interface freezing} follows, where agents establish minimal public APIs in topological order. The {\em implementation} phase sees each node synthesize executable code validated through local contracts. {\em Integration testing} completes the cycle with end-to-end smoke tests that trigger automated critique and repair loops when failures occur.

The syntax and execution feedback loop iteratively refines code until all contracts pass, ensuring every generated repository can successfully train. This graph-native approach enables component-level fault localization, yielding faster convergence compared to monolithic code generation.

\subsubsection*{Stage 4: Visual-Driven Feedback}
The final stage employs visual feedback to iteratively improve implementation quality beyond mere executability. After synthesis produces trainable code, we perform smoke training for 3k iterations (as used in our experiments). The system renders images from multiple camera viewpoints and dispatches them to our critique agent for analysis on three branches. The {\em metric branch} constructs dense error fields by computing local-window PSNR and SSIM maps, identifying regions of highest error through morphological operations. The {\em geometry branch} implements Cross-View Artifact Consensus, highlighting view-inconsistent structures indicative of floaters and ghosting. The {\em semantics branch} leverages Qwen3 VLM to analyze artifact triplets, to output structured diagnoses with candidate patches.

These three feedback mechanisms converge through the master agent, which applies patches within designated regions while maintaining revertability. The framework optionally allows user-driven feedback for domain experts to guide refinement (not used in our experiments for fair evaluations). The refinement loop continues until either: (1) the critique agent produces no further feedback, (2) the maximum iteration count is reached, or (3) the implementation achieves the PSNR targets reported in the original paper. This visual-driven approach consistently converges to within 0.5 dB PSNR and 0.2 SSIM of expert implementations.

\section{Experiments}
\label{sec:experiments}

 This section evaluates \nerfify\ through comprehensive experiments on \nerfifybench\, our specialized benchmark for NeRF paper-to-code synthesis. We first introduce the benchmark composition and baseline systems, then present detailed evaluations of synthesis quality, novelty preservation, and component contributions.

\subsection{\nerfifybench}
\label{sec:Nerfifybench}

To comprehensively evaluate our \nerfify\ framework, we introduce \nerfifybench\, the first specialized benchmark for assessing paper-to-code systems that synthesize trainable NeRF code directly from research papers. While existing benchmarks such as Paper2CodeBench, RexBench, PaperBench, and Code-Dev evaluate general paper-to-code generation, they lack the domain-specific evaluation criteria necessary for NeRF implementations.

\subsubsection{Benchmark Composition}

\nerfifybench\ comprises \textbf{30 carefully selected papers}, classified across four evaluation categories to ensure comprehensive coverage of different implementation challenges:

\noindent \textbf{Set 1.} \textbf{10 Never-Implemented Papers}: Papers without any publicly available source code, with expert-created reference implementations for evaluation. These papers are specifically chosen to avoid LLM training data contamination, since no public implementations exist, the LLMs used in our agents could not have seen corresponding code during pretraining, enabling unbiased evaluation of purely paper-driven synthesis capabilities.\cite{orsingher2023informative,zhu2024vanilla,wang2023anisotropic,sun2024efficient,zhang2024tvnerf,wang2024hyb,deng2023rethinking,yoo2024improving,arandjelovic2021nerf,joung2024stablesurfaceregularizationfast}.

\noindent \textbf{Set 2.} \textbf{5 Non-Nerfstudio Papers}: Papers with existing public implementations but not integrated into the Nerfstudio, allowing direct comparison between generated code and original author implementations. \cite{ma2022deblur,kim2022infonerf,li2024l0,neff2021donerf,garbin2021fastnerf}

\noindent \textbf{Set 3.} \textbf{5 Nerfstudio-Integrated Papers}: Papers already integrated into Nerfstudio, serving as gold-standard references for evaluating synthesis quality. \cite{levy2023seathru,warburg2023nerfbusters,mildenhall2020nerf,muller2022instant,tancik2023nerfstudio}

\noindent \textbf{Set 4.} \textbf{10 Novelty-Coverage Papers}: Papers specifically selected for their distinct technical contributions (novel loss functions, architectural innovations, or training strategies) to evaluate how well our system captures and implements key research innovations. \cite{barron2021mip,passos2024bionerf,turki2023pynerf,chen2022tensorf,kulhanek2023tetra,klenk2023nerf,gu2021stylenerf,yen2021inerf,dihlmann2024signerf,gupta2023mcnerf}


\subsection{Baseline Systems and Evaluation Setup}
\label{sec:baselines}

\noindent \textbf{Baseline Systems.}
We evaluate \nerfify\ against multiple baseline categories spanning generic to specialized approaches. \textit{Paper2Code} and \textit{AutoP2C} represent generic paper-to-code systems that prioritize breadth across all ML domains over domain-specific accuracy. For LLM-based approaches, \textit{GPT-5} (current SOTA) employs single-shot code generation, lacking iterative planning or repairs, while \textit{R1} employs retrieval-augmented in-context learning. We also compare against general multi-agent coding frameworks including \textit{MetaGPT}, \textit{ChatDev}, and \textit{DeepCode}, which operate as software development agents without NeRF-specific architectural understanding or planning. As our gold standard, we include \textit{Expert implementation} by researchers with NeRF expertise. Against these baselines, \textit{Nerfify} employs the complete multi-agent pipeline with Graph-of-Thought synthesis, compositional citation recovery, and closed-loop refinement specifically designed for NeRF implementations.

\noindent \textbf{Computational Setup.}
All experiments are conducted on NVIDIA A6000 GPUs with 48GB memory under consistent configurations. We train scenes for 100k iterations across standard benchmarks including Blender and DTU datasets.

\noindent \textbf{Evaluation Overview.}
Our evaluation examines \nerfify\ from multiple complementary perspectives. Section~\ref{sec:reproduction} compares generated code against expert implementations for papers without public code, measuring visual quality (PSNR, SSIM, LPIPS) vs what they have mentioned in the paper. Section~\ref{sec:novelty} evaluates novelty preservation by analyzing whether systems correctly implement paper-specific innovations including novel equations, loss terms, and architectural components. Section~\ref{sec:ablation} systematically ablates each system component to quantify individual contributions. This comprehensive framework reveals that domain-specific specialization in \nerfify\ dramatically improves both executability (100\% vs 5\% for baselines) and algorithmic fidelity compared to generic approaches.

\subsection{Comparison to Expert Implementations}
\label{sec:reproduction}

\noindent\textbf{Evaluation Metrics.}
We evaluate visual quality using three metrics: PSNR measures pixel-level reconstruction accuracy, SSIM captures perceptual similarity, and LPIPS quantifies perceptual distance using deep features. We report results on standardized test views from paper-specified datasets.

\subsubsection{Set 1: Never-Implemented Papers}
We first compare against expert implementations for NeRF papers without public code, where no code contamination is possible within the LLM's pretrained knowledge, ensuring unbiased assessment of true paper-to-code synthesis capabilities. Expert developers required 1-2 weeks per paper to create reference implementations, while \nerfify\ generates comparable code in minutes. A detailed cost analysis is provided in the supplementary material.

\noindent \textbf{Quantitative Results.}
Table~\ref{tab:comparison} presents quantitative comparisons for representative papers in Set 1. Each paper is evaluated following its paper experiments: KeyNeRF and mi-MLP NeRF report averaged metrics across eight Synthetic-NeRF scenes, ERS uses the DTU dataset, while TVNeRF provides results for a single scene of Synthetic-NeRF(hotdog). \nerfify\ achieves visual quality matching expert implementations within 0.5 dB PSNR and 0.02 SSIM on average. 
Complete results are in the supplementary.

\begin{table}[t]
\centering
\resizebox{\linewidth}{!}{
\begin{tabular}{lccccccccc}
\hline
\textbf{Paper} & \multicolumn{3}{c}{\textbf{Reported}} & \multicolumn{3}{c}{\textbf{Human Impl.}} & \multicolumn{3}{c}{\textbf{\nerfify\ (Ours)}} \\ 
\hline
 & PSNR$\uparrow$ & SSIM$\uparrow$ & LPIPS$\downarrow$ & PSNR$\uparrow$ & SSIM$\uparrow$ & LPIPS$\downarrow$ & PSNR$\uparrow$ & SSIM$\uparrow$ & LPIPS$\downarrow$ \\ 
\hline
KeyNeRF \cite{orsingher2023informative} & 25.65 & 0.89 & 0.11 & 25.70 & 0.89 & 0.12 & 26.12 & 0.90 & 0.09  \\
mi-MLP NerF \cite{zhu2024vanilla} & 24.70 & 0.89 & 0.09 & 22.64 & 0.87 & 0.15 & 22.85 & 0.87 & 0.15  \\
ERS \cite{sun2024efficient} & 27.85 & 0.94 & 0.06 & 26.87 & 0.90 & 0.12 & 27.02 & 0.90 & 0.12 \\
TVNeRF \cite{zhang2024tvnerf} & 27.44 & 0.93 & 0.08  & 26.81 & 0.92 & 0.12 & 27.30 & 0.92 & 0.10 \\
\hline
\end{tabular}}
\vspace{-0.2cm}
\caption{\textbf{Comparison of \nerfify\ with paper and human implementations.} 
We evaluate NeRF papers from the \nerfifybench\ set whose code is not publicly available, using SSIM, PSNR, and LPIPS metrics. \textbf{Note.} Other baselines like Paper2Code, AutoP2C, GPT-5 and R1 failed to generate trainable code.} 
\label{tab:comparison}
\end{table}

In contrast, all baselines often fail to produce executable code as shown in Table~\ref{tab:comparison1}. These failures occur because generic systems cannot recover implicit dependencies from citations or properly structure multi-file architecture. While baselines may generate syntactically valid Python, they lack the domain knowledge to wire components correctly or implement precise mathematical formulations.

\begin{table}[t]
\centering
\resizebox{\columnwidth}{!}{
\begin{tabular}{lccccc}
\toprule
\textbf{Metric} & \textbf{Paper2Code} & \textbf{AutoP2C} & \textbf{GPT-5} & \textbf{R1} & \textbf{Nerfify} \\
\midrule
Imports Resolve & \checkmark & \texttimes & \checkmark & \checkmark & \checkmark \\
Compiles/Trainable & \texttimes & \texttimes & \texttimes & \texttimes & \checkmark \\
Training Stability & \texttimes & \texttimes & \texttimes & \texttimes & \checkmark \\
Converges to Paper Results & \texttimes & \texttimes & \texttimes & \texttimes & \checkmark \\
\bottomrule
\end{tabular}}
\vspace{-0.2cm}
\caption{\textbf{Comparison of \nerfify\ with baselines in terms of executable code.} We evaluate ability to produce functional, trainable implementations. All baselines fail to generate trainable code despite some producing syntactically valid Python.}
\label{tab:comparison1}
\end{table}

\begin{figure}[htbp]
    \centering
    \includegraphics[width=\linewidth]{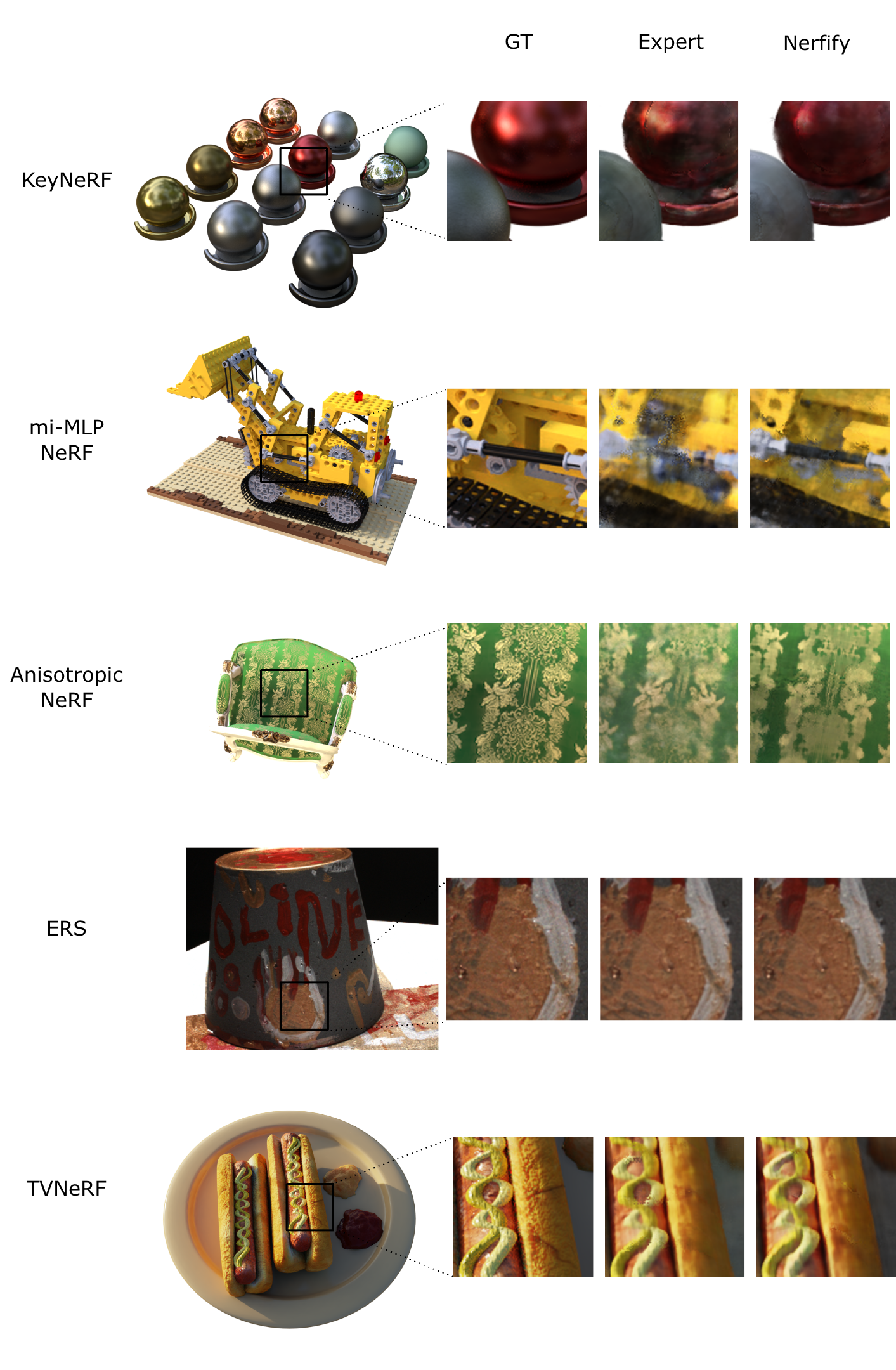}
    \caption{\textbf{Visual Comparison of \nerfify\ and Human Implementation}. \textbf{Left}: Ground Truth Image, \textbf{Middle}: Expert Implementation, \textbf{Right}: Agent Implementation.}
    \label{fig:visual_comparison}
\end{figure}

\noindent\textbf{Qualitative Comparison.}
Figure~\ref{fig:visual_comparison} shows qualitative comparisons on novel viewpoints. \nerfify\ reproduces fine details including specular highlights, geometric edges, and texture patterns, validating that it captures the visual quality and improvements described in papers.

\subsubsection{Sets 2, 3: Papers with Existing Implementations}
Table~\ref{tab:scenario1} shows comparisons between \nerfify\ and existing implementations for four papers from Sets 2 and 3 with public code -- either gold-standard Nerfstudio repositories \cite{mildenhall2020nerf,tancik2023nerfstudio} or original author-provided implementations \cite{li2024l0,kim2022infonerf}. \nerfify\ achieves comparable performance to official implementations with automated Nerfstudio integration. We note that for these papers, LLMs may have encountered their codebases during pretraining. Thereby, \nerfify\ yields exactly the same code as the Nerfstudio repositories for \cite{mildenhall2020nerf,tancik2023nerfstudio}. For papers with non-standard author-provided implementations \cite{li2024l0,kim2022infonerf}, \nerfify\ results in standardized Nerfstudio-compatible code that yields comparable performance. Additional results are shown in supplementary.


\begin{table}[t]
\centering
\resizebox{\columnwidth}{!}{%
\begin{tabular}{lccccccc}
\toprule
\multirow{2}{*}{\textbf{Method}} & \multicolumn{3}{c}{\textbf{Original Repository}} & \multicolumn{3}{c}{\textbf{Nerfify}} \\
\cmidrule(lr){2-4} \cmidrule(lr){5-7}
& PSNR $\uparrow$ & SSIM $\uparrow$ & LPIPS $\downarrow$ & PSNR $\uparrow$ & SSIM $\uparrow$ & LPIPS $\downarrow$ \\
\midrule
Vanilla NeRF~\cite{mildenhall2020nerf} & 31.36 & 0.95 & 0.04 & 31.36 & 0.95 & 0.04 \\
Nerfacto~\cite{tancik2023nerfstudio} & 20.36 & 0.82 & 0.22 & 20.36 & 0.82 & 0.22 \\
$\ell_0$ Sampler~\cite{li2024l0}  & 29.21 & - & 0.04 & 30.13 & 0.97 & 0.03 \\
InfoNeRF~\cite{kim2022infonerf} & 18.27 & 0.81 & 0.23 & 17.87 & 0.69 & 0.44 \\
\bottomrule
\end{tabular}%
}
\vspace{-0.2cm}
\caption{\textbf{Comparison with existing implementations}. Evaluation of \nerfify\ against original author repositories or gold-standard implementations ($\ell_0$ Sampler doesn't report SSIM).}
\label{tab:scenario1}
\end{table}


\begin{table*}[]
\centering
\small
\resizebox{\linewidth}{!}{
\begin{tabular}{l|ccccl|ccccl|ccccl|ccccl|ccccl}
\hline
\multicolumn{1}{c|}{Paper} & \multicolumn{5}{c|}{Paper2Code} & \multicolumn{5}{c|}{AutoP2C} & \multicolumn{5}{c|}{R1} & \multicolumn{5}{c|}{GPT-5} & \multicolumn{5}{c}{Nerfify} \\ \hline
\textbf{} & C$\uparrow$ & I$\downarrow$ & M$\downarrow$ & W$\uparrow$ & Score$\uparrow$ & C$\uparrow$ & I$\downarrow$ & M$\downarrow$ & W$\uparrow$ & Score$\uparrow$ & C$\uparrow$ & I$\downarrow$ & M$\downarrow$ & W$\uparrow$ & Score$\uparrow$ & C$\uparrow$ & I$\downarrow$ & M$\downarrow$ & W$\uparrow$ & Score$\uparrow$ & C$\uparrow$ & I$\downarrow$ & M$\downarrow$ & W$\uparrow$ & Score$\uparrow$ \\ \hline
Mip-NeRF~\cite{barron2021mip} & 0.83 & 0.17 & 0.00 & 0.83 & 0.85 & 0.17 & 0.17 & 0.66 & 0.25 & 0.20 & 0.67 & 0.17 & 0.16 & 0.67 & 0.58 & 0.50 & 0.30 & 0.20 & 0.60 & 0.58 & 1.00 & 0.00 & 0.00 & 1.00 & 1.00 \\
BioNeRF~\cite{passos2024bionerf} & 0.30 & 0.40 & 0.30 & 0.40 & 0.35 & 0.10 & 0.30 & 0.60 & 0.10 & 0.15 & 0.70 & 0.20 & 0.10 & 0.70 & 0.75 & 0.80 & 0.10 & 0.10 & 0.80 & 0.82 & 1.00 & 0.00 & 0.00 & 1.00 & 1.00 \\
PyNeRF~\cite{turki2023pynerf} & 0.50 & 0.30 & 0.20 & 0.60 & 0.58 & 0.00 & 0.10 & 0.90 & 0.10 & 0.03 & 0.30 & 0.60 & 0.10 & 0.70 & 0.68 & 0.40 & 0.30 & 0.30 & 0.80 & 0.52 & 1.00 & 0.00 & 0.00 & 0.90 & 0.97 \\
TensoRF~\cite{chen2022tensorf} & 0.20 & 0.30 & 0.50 & 0.30 & 0.12 & 0.10 & 0.20 & 0.70 & 0.15 & 0.28 & 0.60 & 0.20 & 0.20 & 0.70 & 0.65 & 0.70 & 0.10 & 0.20 & 0.75 & 0.72 & 1.00 & 0.00 & 0.00 & 0.95 & 0.98 \\
Tetra-NeRF~\cite{kulhanek2023tetra} & 0.13 & 0.25 & 0.63 & 0.20 & 0.22 & 0.00 & 0.13 & 0.88 & 0.00 & 0.08 & 0.63 & 0.25 & 0.13 & 0.70 & 0.72 & 0.50 & 0.25 & 0.25 & 0.60 & 0.58 & 1.00 & 0.00 & 0.00 & 1.00 & 1.00 \\
E-NeRF~\cite{klenk2023nerf} & 0.38 & 0.25 & 0.38 & 0.60 & 0.48 & 0.00 & 0.13 & 0.88 & 0.00 & 0.05 & 0.63 & 0.25 & 0.13 & 0.80 & 0.72 & 0.50 & 0.25 & 0.25 & 0.75 & 0.60 & 1.00 & 0.00 & 0.00 & 0.95 & 1.00 \\
StyleNeRF~\cite{gu2021stylenerf} & 0.30 & 0.40 & 0.30 & 0.46 & 0.28 & 0.00 & 0.10 & 0.90 & 0.00 & 0.00 & 0.50 & 0.30 & 0.20 & 0.64 & 0.62 & 0.40 & 0.30 & 0.30 & 0.55 & 0.52 & 1.00 & 0.00 & 0.00 & 1.00 & 0.98 \\
iNeRF~\cite{yen2021inerf} & 0.70 & 0.20 & 0.10 & 0.80 & 0.75 & 0.00 & 0.10 & 0.90 & 0.00 & 0.05 & 0.60 & 0.30 & 0.10 & 0.70 & 0.68 & 0.50 & 0.30 & 0.20 & 0.60 & 0.58 & 1.00 & 0.00 & 0.00 & 1.00 & 0.97 \\
SigNeRF~\cite{dihlmann2024signerf} & 0.38 & 0.38 & 0.24 & 0.50 & 0.52 & 0.00 & 0.13 & 0.87 & 0.00 & 0.08 & 0.63 & 0.25 & 0.12 & 0.75 & 0.72 & 0.50 & 0.25 & 0.25 & 0.63 & 0.58 & 1.00 & 0.00 & 0.00 & 1.00 & 1.00 \\
MCNeRF~\cite{gupta2023mcnerf} & 0.00 & 0.13 & 0.88 & 0.20 & 0.15 & 0.00 & 0.25 & 0.75 & 0.10 & 0.08 & 0.50 & 0.38 & 0.13 & 0.80 & 0.74 & 0.75 & 0.25 & 0.00 & 0.85 & 0.95 & 1.00 & 0.00 & 0.00 & 1.00 & 0.95 \\
\hline
\end{tabular}}
\vspace{-0.2cm}
\caption{\textbf{Novelty coverage analysis across \nerfifybench\ papers}. For each baseline system, we report: C (Correct implementation rate), I (Incorrect/partial implementation rate), M (Missing component rate), W (Hyperparameter weight match accuracy), and Score$_{\text{LLM}}$ (overall semantic implementation score on 0-1 scale). \nerfify\ achieves perfect or near-perfect scores (C=1.00, M=0.00) across all papers, while generic baselines show significant component omissions (M=0.12-0.90 for most methods) and lower implementation fidelity. Metrics are computed over all novel components identified in each paper, with weights derived from paper emphasis and experimental validation.}
\label{tab:my-table}
\end{table*}

\subsection{Novelty Coverage Analysis}
\label{sec:novelty}

We evaluate whether \nerfify\ and baseline methods faithfully implement paper-specific innovations across 10 complex NeRF papers from Set 4 of \nerfifybench\. For each paper, we identify all novelty items including equations, loss terms, architectural blocks, training schedules, etc.

\noindent \textbf{Evaluation.}
Let $\mathcal{N}$ denote the set of novel components identified in paper $P$. We compute four complementary metrics: $C$ measures the fraction of components with {\em correct} implementation, $I$  captures {\em incomplete} but partially correct implementations, $M$ for {\em missing} unimplemented components, and $W = |\{n \in \mathcal{N} : |\theta_n - \hat{\theta}_n| < 0.1|\theta_n|\}|/|\mathcal{N}|$ evaluates fidelity of {\em weights} where $\theta_n$ and $\hat{\theta}_n$ represent paper-specified and implemented values, respectively. 
Note that \nerfify\ can adaptively modify $\hat{\theta}_n$ through its Visual-Driven Feedback to achieve better training dynamics than the original paper specifications, distinguishing it from baselines that attempt only exact reproduction.
To capture semantic understanding beyond structural metrics, we employ an LLM-based {\em score} that assesses implementation fidelity as $\left( {\sum_{i=1}^{n} w_i \cdot s_i} \right)/{\sum_{i=1}^{n} w_i}$,
where $w_i \in [0,1]$ represents the importance of novelty $i$ extracted from paper, and $s_i \in \{0, 0.2, 0.4, 0.6, 0.8, 1.0\}$ indicates implementation completeness on a 6-level scale ranging from unimplemented to fully correct (details in supplementary).
The LLM score analyzes both mathematical formulations and algorithmic logic, recognizing equivalent implementations and scoring each novelty component and its weight.

\noindent \textbf{Results.}
As shown in Table~\ref{tab:my-table}, \nerfify\ consistently achieves higher correct implementation rates across all evaluated papers compared to baselines. \nerfify\ achieves perfect or near-perfect scores (C=1.00, M=0.00) across all papers, while generic baselines show significant component omissions and lower implementation fidelity. We note that achieving competent code generation from baseline LLM-based systems typically requires expert-level prompt engineering, which \nerfify\ does not require.
Complete paper-to-code snippet comparisons for all identified novelties and additional comparison with code-generation tools like ChatDev \cite{qian2023chatdev}, MetaGPT \cite{hong2023metagpt}, and DeepCode \cite{deepcode2025} are provided in the supplementary material.

\begin{table}[t]
\centering
\resizebox{\columnwidth}{!}{
\begin{tabular}{lccc}
\toprule
\textbf{Configuration} & \textbf{Score} & \textbf{Trainable} & \textbf{Correct Novelties} \\
 & & \textbf{(\%)} & \textbf{Impl. (C)} \\
\midrule
\nerfify\ (Full) & 0.98 & 100 & 1.00 \\
\midrule
\textit{Knowledge Sources:} \\
\quad w/o In-context Examples (Stage 1) & 0.71 & 90 & 1.00 \\
\quad w/o citation recovery (Stage 2) & 0.68 & 100 & 0.65 \\
\quad w/o Both & 0.58 & 90 & 0.65 \\
\midrule
\textit{Validation \& Feedback:} \\
\quad w/o Smoke Tests (Stage 3) & 0.69 & 60 & 0.85 \\
\quad w/o VLM Feedback (Stage 4) & 0.99 & 100 & 1.00 \\
\midrule
\textit{Planning Strategy:} \\
\quad One-Shot (no GoT) (Stage 3) & 0.45 & 70 & 1.00 \\
\bottomrule
\end{tabular}}
\vspace{-0.2cm}
\caption{\textbf{Component ablation study}. We evaluate the impact of each system component on synthesis quality and efficiency. Numbers are averaged over 10 papers from \nerfifybench.}
\label{tab:ablation_all}
\end{table}

\subsection{Ablation Study}
\label{sec:ablation}
In Table~\ref{tab:ablation_all}, we selectively disable key elements of \nerfify\ (other metrics like PSNR are in supplementary). The results validate our multi-agent architecture, demonstrating the importance of domain knowledge, compositional reasoning, iterative validation, and visual refinement.


\noindent \textbf{Knowledge Sources.}
Removing in-context examples drops score from 0.98 to 0.71 while trainability falls to 90\%, though novelty implementation remains perfect (C=1.00). This indicates agents can interpret equations correctly but struggle with architectural integration. Citation dependencies proves equally critical: without it, the system maintains trainability but fails to implement 35\% of novel techniques (C=0.65). Disabling both reduces semantic score to 0.58 while trainability remains at 90\%, confirming that code quality degradation stems primarily from improper architectural patterns rather than runtime failures. Ablation on $|\mathcal{K}|$ in the supplementary material.

\noindent \textbf{Validation and Feedback.}
Smoke tests significantly impact trainability. Without incremental validation, trainability drops to 60\% and correct novelty implementation falls to 0.85 as interface mismatches accumulate during generation. VLM-guided feedback maintains perfect implementation (C=1.00) but slightly reduces semantic score from 0.99 to 0.98, reflecting hyperparameter adjustments that prioritize practical convergence over strict paper fidelity.

\noindent \textbf{Planning Strategy.}
GoT synthesis substantially improves code quality over one-shot generation. While one-shot maintains perfect novelty implementation (C=1.00) and 70\% trainability, its semantic score collapses to 0.45, indicating failures in establishing proper module boundaries and abstractions despite correct equation implementation.

\section{Conclusion}
We presented \nerfify, a multi-agent framework that enables paper-to-code synthesis for fully trainable NeRF implementations through domain-aware reasoning and structured code generation. By formalizing Nerfstudio as a context-free grammar and orchestrating code generation through graph-of-thought planning, the framework ensures architectural correctness and full executability across complex NeRF pipelines. Its compositional citation recovery and visual-driven feedback enable agents to reconstruct and refine hidden dependencies, achieving expert-level visual quality within 0.5 dB PSNR while reducing development time from weeks to minutes. Beyond accelerating NeRF research, \nerfify\ highlights how depth of specialization, rather than model scale alone, enables reliable translation of scientific ideas into working code. This insight suggests a template where domain-specific grammars and agentic reasoning can transform research reproducibility in other communities too. Future work will extend \nerfify\ to further NeRF variants and NeRF-based methods, other areas of computer vision research and broader paper-to-experiment frameworks.



{
    \small
    \bibliographystyle{ieeenat_fullname}
    \bibliography{main}
}

\end{document}